# RaFoLa: A Rationale-Annotated Corpus for Detecting Indicators of Forced Labour

**Erick Mendez Guzman, Viktor Schlegel and Riza Batista-Navarro**
Department of Computer Science, University of Manchester
Manchester, United Kingdom
{erick.mendezguzman, viktor.schlegel, riza.batista}@manchester.ac.uk

**Abstract**

Forced labour is the most common type of modern slavery, and it is increasingly gaining the attention of the research and social community. Recent studies suggest that artificial intelligence (AI) holds immense potential for augmenting anti-slavery action. However, AI tools need to be developed transparently in cooperation with different stakeholders. Such tools are contingent on the availability and access to domain-specific data, which are scarce due to the near-invisible nature of forced labour. To the best of our knowledge, this paper presents the first openly accessible English corpus annotated for multi-class and multi-label forced labour detection. The corpus consists of 989 news articles retrieved from specialised data sources and annotated according to risk indicators defined by the International Labour Organization (ILO). Each news article was annotated for two aspects: (1) indicators of forced labour as classification labels and (2) snippets of the text that justify labelling decisions. We hope that our data set can help promote research on explainability for multi-class and multi-label text classification. In this work, we explain our process for collecting the data underpinning the proposed corpus, describe our annotation guidelines and present some statistical analysis of its content. Finally, we summarise the results of baseline experiments based on different variants of the Bidirectional Encoder Representation from Transformer (BERT) model.

## 1. Introduction

Forced labour is the most common type of modern slavery, affecting at least 24.9 million people worldwide (Landman and Silverman, 2019). The trail of forced labour and the data that could reveal its international network is spread across many organisational and geographical boundaries (Pasley, 2018).

Recent evidence suggests that AI can facilitate efforts to combat modern slavery. For instance, network analysis and anomaly detection have been utilised to identify patterns in financial flows as well as to detect populations targeted for exploitation (Milivojevic et al., 2020; Bliss et al., 2021). Furthermore, AI models have been used to analyse mobile phone data, specifically mobile money, to predict the incidence of forced labour in Africa (Milivojevic et al., 2020). Considering that most of the data regarding modern slavery come in the form of text records, more recent efforts have focused on using Natural Language Processing (NLP) methods to boost the detection of forced labour (Pasley, 2018).

Eliminating modern slavery and forced labour has been particularly challenging due to its near-invisible nature (Landman and Silverman, 2019). Nevertheless, NLP methods can identify individuals and patterns from unstructured data to detect possible red flags that indicate exploitation (Bliss et al., 2021). To accomplish this, NLP models must be anchored to human rights considerations and developed in cooperation with different stakeholders (Milivojevic et al., 2020).

The increasing deployment of AI tools in high-stake domains has been coupled with increased societal demands for these systems to explain their predictions (Arrieta et al., 2020). Consequently, Explainable Artificial Intelligence (XAI) has emerged as a research field aiming to develop methods and techniques that allow human users to understand outcomes produced by AI systems (Došilović et al., 2018).

Text classification is a fundamental task in the field of Natural Language Processing (NLP), whereby predefined categories are automatically assigned to free-text documents (Vijayan et al., 2017). Text classifiers are an essential component in many NLP applications such as web searching, information retrieval and sentiment analysis, among many others (Aggarwal and Zhai, 2012). While deep learning-based methods have substantially improved model accuracy for text classification, they come at the expense of becoming less interpretable (Danilevsky et al., 2020). Understanding the inner workings of a text classifier is challenging, considering not just model complexity but also the size of the documents and the variety of tokens involved in the classification problem (Arrieta et al., 2020).

Explanations in NLP often take the form of *rationales* (Figure 1), defined as a subset of input tokens that are considered relevant to a model's decision (Lei et al., 2016). A rationale should be a short yet sufficient part of the input text (DeYoung et al., 2019): short so that it makes clear what the most important part of the input sequence is, and sufficient so that the correct prediction

Moreover, a section of mid-level officers of some garment factories hurl <mark style="background-color:#8888cc">abusive words at female workers and even go for sexual harassment</mark>.

<mark style="background-color:#88cc99">If a female worker protests such kind of attitude, she is either terminated or faces a reverse complaint</mark>, Shanta said.

Figure 1: Example of rationales supporting the identification of forced labour indicators, shown in different colours, within the text of a news article.

can be made from the rationale alone (Bastings et al., 2019). Similarly, *human rationales* are snippets of text input marked by human annotators to justify their labelling decisions. Recent evidence suggests that humans providing explanations not only boost the accuracy of machine learning-based models but also improve the quality of their explanations (Strout et al., 2019).

Given that a single input text might describe multiple forced labour-related violations, we focus our work on multi-class and multi-label text classifiers for identifying forced labour indicators and attempt to make their predictions more understandable. Our goal is to provide richer annotations for training text classification models, i.e., labels with rationales. When annotating a news article, our annotators also highlight the evidence supporting their annotation, thereby allowing classifiers to learn *why* the instance belongs to a certain category.

To summarise, our main contributions in this paper are as follows:

- We design a rationale-oriented annotation scheme for capturing indicators of forced labour within news articles.

- To the best of our knowledge, we present the first resource consisting of news articles annotated for indicators of forced labour, and their respective human-generated rationales.

- We provide results of multi-class and multi-label baseline models to predict such indicators.

The rest of the paper is organised as follows: Section 2 describes our data collection process, annotation schema and results of the annotation task. Section 3 presents our baseline experiments carried out on the proposed corpus. Section 4 discusses prior NLP work related to modern slavery. We conclude by discussing our work and providing perspectives for future research in Section 5.

## 2. Data Collection and Annotation

This section describes the process for collecting the documents constituting our corpus and their annotation process. The annotated data set has been released under the Creative Commons Attribution-NonCommercial 4.0 International License (CC-BY-NC-4.0)[1].

### 2.1. Data collection and pre-processing

Even though the problem of modern slavery has increased in terms of relevance and awareness since the 1990s, there is not much information about it in general news outlets (Lucas and Landman, 2021). Therefore, the first step is to identify sources containing data on forced labour specifically. Consequently, we collect news articles from the following data sources:

- **Traffik Analysis Hub** (TAH, 2012): A partnership program across industries and sectors to share data regarding modern slavery and human trafficking. Registration is compulsory for using the platform. We obtained access to the platform by registering as an academic institution.

- **Business & Human Rights Resource** (BHR, 2015): A research organisation dedicated to advancing human rights in business and eradicating abuse. News articles are freely available on the website.

- **Internation Labour Organization Newsroom** (ILO, 2020): The International Labour Organization (ILO) is a United Nations agency aiming to advance social and economic justice by establishing international labour standards. The newsroom web pages are publicly accessible.

A common feature across all these data sources is that they operate as repositories that store news articles reporting different human rights violations across the globe. We retrieved the URLs of news articles which are: written in English, posted from January 2019 to September 2021, and categorized under the labels of 'modern slavery/forced labour', 'forced labour', and 'labour exploitation'. As these platforms are independent, there is some redundancy among the retrieved news articles; to alleviate this, we removed duplicated URLs.

Considering that news articles are published across media outlets globally with varied HTML schemata, we made use of the Diffbot (Diffbot, 2018) web scraping tool to extract the title and content automatically from each news article. After consolidating the news articles into a combined data set, we removed items with identical titles. Finally, a data set consisting of 989 news articles was annotated according to our guidelines.

---
[1] https://creativecommons.org/licenses/by-nc/4.0/legalcode

## 2.2. Annotation guidelines

Before performing the annotation, we developed an annotation scheme to guide annotators in labelling news articles. The scheme is based on the 11 indicators of forced labour defined by the ILO (ILO, 2012), as suggested by our project advisors consisting of domain experts from academia and NGOs. These indicators are intended to help law enforcement officials, labour inspectors, and NGO workers to identify persons who are possibly trapped in a forced labour situation. For a detailed description of the indicators of forced labour and examples for each one of them, we refer the reader to Appendix A.

As mentioned in the previous section, we focussed on multi-class and multi-label text classification. Therefore, the annotation process assigns one or more of the forced labour indicators to each news article. However, since our goal is to provide richer annotations that support text classification, we asked our annotators to select phrases and sentences that support their labelling decisions. For the annotation guidelines, we refer the reader to Appendix A.

## 2.3. Annotation task

The annotation of the corpus was completed by two annotators, a male (A1) and a female (A2), both adults aged over 30 with Master-level degrees from the United Kingdom. Considering our domain of interest and aim, we decided against crowd-sourcing the annotation task to allow for working directly with the annotators and exchanging qualitative feedback with them, and to ensure high-quality annotation of the rationales (Nowak and Rüger, 2010).

We first randomly selected 100 news articles and asked our annotators to annotate them independently using LightTag (LightTag, 2018) as the annotation tool. This preliminary task helped the annotators to familiarise themselves with the topic and to understand the scope of the task. More importantly, this enabled us to obtain constructive feedback on the annotation guidelines.

Inter-annotator agreement (IAA) is commonly calculated to assess the quality of annotations in corpus linguistics (Krippendorff, 2004). Since our annotation is a two-fold task, we computed IAA metrics at the level of both labels and rationales. Considering that many researchers have utilised the F1 score for multi-label settings, we report IAA for labels by calculating the micro-averaged F1 score (Nowak and Rüger, 2010). Considering A1's annotations as the gold standard, the overall F1 score for these 100 news articles is 0.81, meaning that on average, nearly two-thirds of the total labels were agreed on by both annotators. For a per-class breakdown of IAA results, we refer the reader to Appendix B.

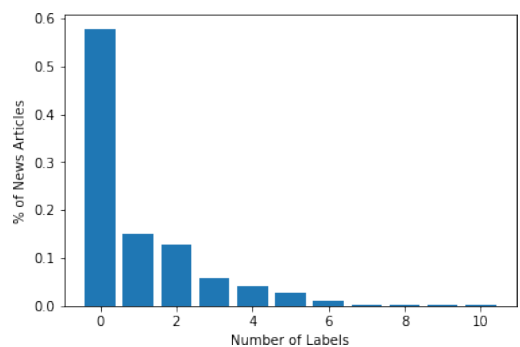

Figure 2: Distribution of the number of labels

Measuring exact matches between human-generated rationales is likely to be too strict. Consequently, for calculating IAA for rationales, we used the Intersection-Over-Union (IOU) at the token level (DeYoung et al., 2019). For two human-generated rationales, IOU is the size of the overlap of the tokens covered, divided by the size of their union. The rationales from two annotators are counted as a match if the overlap between them is more than a threshold, which is 0.5 for our study following Zaidan et al. (2007). Finally, we use these partial matches to calculate a micro-averaged F1 score for rationales of 0.73.

We observed a fair agreement between both annotators and consider these IAA metrics satisfactory given the novelty of our data set and the complexity of the annotation task. Each of the remaining news articles were subsequently annotated by one annotator (600 and 289 articles by A1 and A2, respectively).

## 2.4. Annotation results

In this section, we report descriptive statistics based on the annotated data set. Overall, the corpus is comprised of 989 news articles and 5,026,746 words, out of which 36,386 are unique. The news articles are lengthy documents, having 4,957 words on average and a standard deviation of 4,516 words.

Figure 2 illustrates the number of forced labour indicators assigned to each news article within the corpus. On average, each news article is assigned or tagged with 1.2 labels (forced labour indicators). However, only 43% of them were tagged with at least one label. Even though articles were drawn from specialised data sources under specific categories, many of them describe forced labour cases in general without referring to any ILO indicators in particular. An example of this is a news article describing the increase in the number of cases of forced labour in a specific region, country or sector but without describing any abusive practice in detail.

Table 1 provides an overview of the distribution of the forced labour indicators across the annotated articles,

| Forced Labour Indicator | # News Articles | Frequent Words (Articles) | Frequent Words (Rationales) |
|---|---|---|---|
| Abuse of vulnerability | 172 | [*workers, labour, work, rights, forced*] | [*vulnerable, child, children, forced, women*] |
| Abusive working and living conditions | 256 | [*workers, rights, children, human, palm*] | [*conditions, water, living, food, dangerous*] |
| Debt bondage | 72 | [*workers, labour, migrant, trafficking, human*] | [*pay, debt, fees, money, recruitment*] |
| Deception | 51 | [*workers, trafficking, labour, slavery, victims*] | [*promise, job, lured, contracts, recruitment*] |
| Excessive overtime | 117 | [*workers, labour, palm, oil, children*] | [*hours, day, work, week, plantation*] |
| Intimidation and threats | 67 | [*workers, women, labour, forced, rights*] | [*threats, retaliation, refused, reported, bosses*] |
| Isolation | 47 | [*palm, oil, workers, children, plantations*] | [*plantations, remote, phone, guarded, hills*] |
| Physical and sexual violence | 123 | [*workers, labour, children, women, forced*] | [*abuse, sexual, harassment, violence, beaten*] |
| Restriction of movement | 34 | [*workers, labour, forced, trafficking, conditions*] | [*locked, factory, guard, armed, escaping*] |
| Retention of identity documents | 31 | [*workers, labour, trafficking, force, human*] | [*passport, documents, taken, confiscated, migrant*] |
| Withholding of wages | 47 | [*workers, labour, rights, people, human*] | [*wages, pay, unpaid, withheld, money*] |

Table 1: Number of news articles and most frequently occurring words for each forced labour indicator

alongside the most frequently occurring words in corresponding text and rationales. Among the 425 news articles that contain at least one forced labour indicator, 60.2% of them are labelled as 'Abusive working and living conditions'. Furthermore, 40.4% and 28.9% are tagged as 'Abuse of vulnerability' and 'Physical and sexual violence', respectively.

As shown in Table 1, no significant differences were found among the most occurring words in news articles when compared across indicators of forced labour. Words like 'workers' and 'labour' are frequently found regardless of the label in the news article. In contrast, rationales exhibit a distinctive pattern for each indicator. This is an interesting result, indicating that human-annotated rationales might carry relevant information towards a model's decision.

Many real-world classification tasks, including our own, require handling of highly unbalanced data sets, in which the number of samples from one class is much smaller than that from other classes (Tahir et al., 2012). It is essential to note that the class imbalance problem is an inherent characteristic of multi-label data, which hinders both the accuracy and explainability of most learning methods (Danilevsky et al., 2020).

## 3. Experiments and Results

We conduct experiments using state-of-the-art NLP models based on pre-trained language models such as BERT, to establish a baseline for multi-class and multi-label classification on our data set[2].

### 3.1. Classifiers

Following Devlin et al. (2018), we represent each English news article in its raw text form and insert the special `[CLS]` token at the beginning of the sequence. The text is embedded using the language model, and the embedding of the `[CLS]` token is projected into an eleven-dimensional space. We minimise the binary cross-entropy loss during training between the logits (the unnormalised model predictions) and the expected labels. Finally, we pass each logit through a sigmoid function as the model's predictions for each label for inference.

For the task at hand, we fine-tuned the following transformer-based models on our data set:

- **DistilBERT** (Sanh et al., 2019): A smaller and faster transformer model trained by distilling BERT base (Devlin et al., 2018).

- **ALBERT** (Lan et al., 2019): A light version of BERT (Devlin et al., 2018) that uses parameter-reduction techniques that allow for large-scale configurations.

- **RoBERTa** (Liu et al., 2019): A retraining of BERT with improved architecture and training

---

[2]Code can be retrieved from https://github.com/emendezguzman/rationales_forced_labour

methodology. For this model, we use the 'base', 'distil-roberta' and 'large' versions.

- **XLNet** (Yang et al., 2019): A generalized autoregressive pre-trained method that uses improved training methodology and larger data than BERT (Devlin et al., 2018).

Since there is a relatively small body of literature on using state-of-the-art NLP methods in the humanitarian domain, we decided to explore and utilise a set of BERT variations considering the trade-off between performance and computational cost.

On the one hand, we selected DistilBERT and ALBERT based on the basis of their faster training time and inference. While DistilBERT learns a distilled version of BERT that retains 97% performance while using only half the number of parameters (Sanh et al., 2019), ALBERT introduces architecture and training changes to reduce the model size (Lan et al., 2019).

On the other hand, data from several studies suggest that RoBERTa and XLNet outperform BERT variations on benchmark results (Adoma et al., 2020; Cortiz, 2021). XLNet introduces permutation language modelling, helping the model better handle dependencies and relations between words (Yang et al., 2019). Finally, RoBERTa removes Next Sentence Prediction from BERT's pretraining and introduces dynamic masking to achieve better performance (Liu et al., 2019).

### 3.2. Experimental setup

The classification experiments were performed with the aid of the Simple Transformers library (Rajapakse, 2019). Simple Transformers is a Python package based on the Transformers library by HuggingFace (Wolf et al., 2019), which was designed to simplify the usage of transformer models whilst preserving their architecture.

The performance of these state-of-the-art models depends not only on the parameter values that the model learns during training but also on the values of their hyperparameters (Devlin et al., 2018). Thus, we split the data set into training, validation and test sets according to a 70:10:20 ratio and search for the hyperparameter values that minimise the function loss over the validation set.

To optimise the training process, we tuned the model hyperparameters using a random search method (Bergstra and Bengio, 2012) and run a total of 40 training runs, one for each combination of hyperparameters. Each trial was fine-tuned for three epochs on the training set. For a detailed description of the hyperparameter tuning process and its results, we refer the reader to Appendix C.

Finally, we merged the training and validation sets in preparation for fine-tuning the models. The classifiers were trained for ten epochs using the hyperparameters selected by the search method described above. The classifiers' performance were then evaluated on the test set.

Considering that the classes in our annotated corpus are highly imbalanced and that there are many articles without any forced labour indicators, we decided to apply a simple random under-sampling method over the training and validation sets. Consequently, we ran our experiments on the following data sets (Tahir et al., 2012):

- **Data set 1**: The whole corpus, including the news articles without any assigned labels ($n$=989).

- **Data set 2**: We removed half of the news articles without any assigned labels ($n$=763 which were randomly selected).

- **Data set 3**: We kept only news articles with at least one label assigned ($n$=538).

### 3.3. Performance metrics

We employed three metrics to evaluate the performance of our baseline classifiers: F1 Score (F1), Label Ranking Precision Average Precision Score (LRAP), and Exact Match Ratio (EMR) (Feldman et al., 2007).

F1 is the harmonic mean of precision and recall. As there are multiple ways to obtain a single F1 score indicator for multi-class and multi-label classification, we decided to utilise the micro-averaged (micro), macro-averaged (macro), and weighted F1 scores (Feldman et al., 2007). The weighted F1 score, as the average weighted by the number of true instances for each label, takes into account the class imbalance in our corpus (Ghamrawi and McCallum, 2005).

LRAP is a metric used for multi-label classification problems that, for each ground truth label, evaluate the fraction of higher-ranked labels that were correctly predicted (Schapire and Singer, 2000). It is important to note that LRAP is a threshold-independent metric scoring between 0 and 1, with 1 being the best value. Finally, EMR computes the proportion of labels predicted by a model that matches the corresponding set of ground truth labels (Ghamrawi and McCallum, 2005). A disadvantage of this measure is that it does not distinguish between perfect and partially incorrect matches (Feldman et al., 2007).

### 3.4. Results

The results obtained for each classifier on the test set of Data set 1 (original corpus) are presented in Table 2.

What stands out in the table are the models' LRAP scores, all of them being equal to or greater than 0.85,

| Model | F1(weighted) | F1(micro) | F1(macro) | LRAP | EMR |
|---|---|---|---|---|---|
| roberta-base | 0.47 | 0.45 | 0.41 | 0.86 | 0.49 |
| distilroberta-base | 0.49 | 0.49 | 0.45 | 0.85 | 0.50 |
| distilbert-base | 0.49 | 0.48 | 0.47 | 0.86 | 0.48 |
| xlnet-base | 0.51 | 0.52 | 0.47 | 0.87 | 0.51 |
| albert-base | 0.47 | 0.47 | 0.46 | 0.86 | 0.44 |
| roberta-large | 0.47 | 0.47 | 0.44 | 0.86 | 0.55 |

Table 2: Results on the test subset of Data set 1

| Model | Dataset | F1(weighted) | LRAP | EMR |
|---|---|---|---|---|
| roberta-base | Data set 1 | 0.47 | 0.86 | 0.49 |
| | Data set 2 | 0.45 | 0.87 | 0.42 |
| | Data set 3 | 0.40 | 0.88 | 0.05 |
| distilroberta-base | Data set 1 | 0.49 | 0.85 | 0.50 |
| | Data set 2 | 0.50 | 0.88 | 0.43 |
| | Data set 3 | 0.43 | 0.89 | 0.09 |
| distilbert-base | Data set 1 | 0.49 | 0.86 | 0.48 |
| | Data set 2 | 0.44 | 0.88 | 0.25 |
| | Data set 3 | 0.36 | 0.88 | 0.06 |
| xlnet-base | Data set 1 | 0.51 | 0.87 | 0.51 |
| | Data set 2 | 0.44 | 0.86 | 0.43 |
| | Data set 3 | 0.38 | 0.87 | 0.06 |
| albert-base | Data set 1 | 0.47 | 0.86 | 0.44 |
| | Data set 2 | 0.47 | 0.88 | 0.34 |
| | Data set 3 | 0.35 | 0.87 | 0.04 |
| roberta-large | Data set 1 | 0.47 | 0.86 | 0.55 |
| | Data set 2 | 0.46 | 0.87 | 0.49 |
| | Data set 3 | 0.39 | 0.88 | 0.13 |

Table 3: Results on the test subsets of our three data sets

meaning that models assign a higher probability to the truly positive labels. In terms of weighted F1 score, however, the models have some room for improvement when compared with previous research in multi-label classification on news articles (Madjarov et al., 2012). Interestingly, the XLNet records the highest micro, macro, and weighted F1 scores with 0.52, 0.47, and 0.51, respectively. It is important to note that XLNet is the only auto-regressive method among all tested models suggesting that the model's permutation-based training helps in improving its performance on our data set.

One unexpected finding concerning RoBERTa is that an increase in the model's size does not necessarily imply better performance. As shown in Table 2, roberta-large does not outperform significantly smaller versions of the same architecture, namely roberta-base and distilroberta-base. These results, however, might be affected by our limited sample size; more experiments are needed to derive more robust conclusions (Madjarov et al., 2012).

Table 3 compares the results for each model across our data sets. This table is revealing in several ways. First, the weighted F1 score decreases as we undersample examples with no labels. Almost all models, except for the distilroberta-base, worsened their F1 scores compared to their results on Data set 1. These results do not align with the findings of many previous efforts on sampling methods for addressing the class imbalance problem (Tahir et al., 2012).

Second, there is a clear trend of decreasing EMR scores when removing examples without labels. A possible explanation for this might be that the models overfit to instances with no labels due to class imbalance. Even though the LRAP score remains relatively stable across data sets, there is an impact on the model's capacity to match ground-truth labels. To illustrate, the EMR drops, on average, to a fifth of its value when comparing results on Data sets 1 and 3.

Finally, Table 4 shows the per-class F1 scores for the best performing XLNet model across our data sets. Even though the model performs comparatively well for some labels, for instance, 'Abusive working and living conditions'. Note that results are lower for under-represented classes such as 'Intimidation and threats', 'Retention of identity documents' and 'Withholding of wages', which clearly leaves some room for improvement. For a per-class breakdown of results for each classifier and data set, we refer the reader to Appendix D.

| Label | Data set 1 | Data set 2 | Data set 3 |
|---|---|---|---|
| Abuse of vulnerability | 0.34 | 0.49 | 0.59 |
| Abusive working and living conditions | 0.68 | 0.66 | 0.79 |
| Debt bondage | 0.50 | 0.40 | 0.40 |
| Deception | 0.50 | 0.50 | 0.36 |
| Excessive overtime | 0.58 | 0.30 | 0.50 |
| Intimidation and threats | 0.20 | 0.61 | 0.40 |
| Isolation | 0.55 | 0.50 | 0.44 |
| Physical and sexual violence | 0.38 | 0.52 | 0.53 |
| Restriction of movement | 0.75 | 0.44 | 0.28 |
| Retention of identity documents | 0.22 | 0.28 | 0.00 |
| Withholding of wages | 0.54 | 0.40 | 0.00 |

Table 4: Results of the best performing XLNet model on the test subset

Overall, we observe that while the corpus is well-suited for the multi-class and multi-label setting, the classification task is not easily solved, even with the use of state-of-the-art transformer-based methods.

## 4. Related Work

Only a relatively small body of literature is concerned with applying NLP methods to the humanitarian domain. To the best of our knowledge, the only publicly available resource is a corpus of Arabic tweets developed to support the automatic identification of human rights abuses (Alhelbawy et al., 2016). Even though there is no record of previous research conducted on modern slavery or forced labour, some studies attempted to use sentiment analysis to identify human rights violations. For this purpose, researchers have utilised state-of-the-art deep learning techniques to detect human rights abuses as a binary classification task either on social media platforms (Alhelbawy et al., 2020) or messaging applications (Nomnga and Ngqulu, 2021).

## 5. Conclusion and Future Work

NLP tools hold immense potential for supporting anti-slavery action. Deep learning models can help identify trends from text data, facilitate effective early detection, and identify individuals at risk of being victims of modern slavery. However, the unavailability of annotated domain-specific data has been a significant setback. To bridge this knowledge gap, we introduce a rationale-annotated corpus focussed on forced labour, which will support the development of models for multi-class and multi-label text classification. The novelty of our data set is that news articles have been labelled with both indicators of forced labour and word-level rationales that support labelling decisions.

Furthermore, we have presented a set of text classifiers for detecting indicators of forced labour using transformer-based models that can serve as a strong baseline for future work in this direction. Even though our sample size may somewhat limit the findings, results provide an attractive starting point for researchers interested in text classification in the humanitarian domain.

We seek to establish whether human-generated rationales can aid learning and explainability. Our future work aims at incorporating them during training to improve a model's predictive performance, and the quality of its generated explanations (Strout et al., 2019; Lei et al., 2016). Finally, we hope that our data set can help promote research on explainability in NLP, specifically for multi-class and multi-label text classification problems.

## 6. Acknowledgements

We wish to acknowledge and thank Quintin Lake (Fifty Eight), Jimena Monjaras (Contratados) and Malte Skov (Global Development Institute at The University of Manchester) for their advice on modern slavery and the ILO Indicators of Forced Labour.

## 7. Bibliographical References

Adoma, A. F., Henry, N.-M., and Chen, W. (2020). Comparative Analyses of BERT, RoBERTa, Distil-BETR, and XLNet for Text-Based Emotion Recognition. In *2020 17th International Computer Conference on Wavelet Active Media Technology and Information Processing (ICCWAMTIP)*, pages 117–121. IEEE.

Aggarwal, C. C. and Zhai, C. (2012). A Survey of Text Classification Algorithms. In *Mining text data*, pages 163–222. Springer.

Alhelbawy, A., Massimo, P., and Kruschwitz, U. (2016). Towards a corpus of violence acts in Arabic social media. In *Proceedings of the Tenth International Conference on Language Resources and Evaluation (LREC'16)*, pages 1627–1631.

Alhelbawy, A., Lattimer, M., Kruschwitz, U., Fox, C., and Poesio, M. (2020). An NLP-Powered Human

Rights Monitoring Platform. *Expert Systems with Applications*, 153:113365.

Arrieta, A. B., Diaz-Rodriguez, N., Del Ser, J., Bennetot, A., Tabik, S., Barbado, A., Garcia, S., Gil-Lopez, S., Molina, D., and Benjamins, R. (2020). Explainable Artificial Intelligence (XAI): Concepts, taxonomies, opportunities and challenges toward responsible AI. *Information Fusion*, 58:82–115.

Bastings, J., Aziz, W., and Titov, I. (2019). Interpretable Neural Predictions with Differentiable Binary Variables. *arXiv preprint arXiv:1905.08160*.

Bergstra, J. and Bengio, Y. (2012). Random Search for Hyper-Parameter Optimization. *Journal of machine learning research*, 13(2).

BHR. (2015). Business & human rights resource centre. https://www.business-humanrights.org/en/.

Bliss, N., Briers, M., Eckstein, A., Goulding, J., Lopresti, D., Mazumder, A., and Smith, G. (2021). CCC/Code 8.7: Applying AI in the Fight Against Modern Slavery. *arXiv preprint arXiv:2106.13186*.

Cortiz, D. (2021). Exploring Transformers in Emotion Recognition: A Comparison of BERT, DistillBERT, RoBERTa, XLNet and ELECTRA. *arXiv preprint arXiv:2104.02041*.

Danilevsky, M., Qian, K., Aharonov, R., Katsis, Y., Kawas, B., and Sen, P. (2020). A Survey of the State of Explainable AI for Natural Language Processing. *arXiv preprint arXiv:2010.00711*.

Devlin, J., Chang, M.-W., Lee, K., and Toutanova, K. (2018). BERT: Pre-training of Deep Bidirectional Transformers for Language Understanding. *arXiv preprint arXiv:1810.04805*.

DeYoung, J., Jain, S., Rajani, N. F., Lehman, E., Xiong, C., Socher, R., and Wallace, B. C. (2019). ERASER: A Benchmark to Evaluate Rationalized NLP Models. *arXiv preprint arXiv:1911.03429*.

Diffbot. (2018). Knowledge graph, ai web data extraction and crawling. https://www.diffbot.com/.

Došilović, F. K., Brčić, M., and Hlupić, N. (2018). Explainable Artificial Intelligence: A Survey. In *2018 41st International convention on information and communication technology, electronics and microelectronics (MIPRO)*, pages 0210–0215. IEEE.

Feldman, R., Sanger, J., et al. (2007). *The Text Mining Handbook: Advanced Approaches in Analyzing Unstructured Data*. Cambridge university press.

Ghamrawi, N. and McCallum, A. (2005). Collective Multi-label Classification. In *Proceedings of the 14th ACM international conference on Information and knowledge management*, pages 195–200.

ILO. (2012). ILO Indicators of Forced Labour. In: Special Action Programme to Combat Forced Labour (SAP-FL). *Special Action Programme to Combat Forced Labour*.

ILO. (2020). Ilo newsroom. https://www.ilo.org/global/about-the-ilo/newsroom/lang--en/index.htm.

Krippendorff, K. (2004). Measuring the Reliability of Qualitative Text Analysis Data. *Quality and Quantity*, 38:787–800.

Lan, Z., Chen, M., Goodman, S., Gimpel, K., Sharma, P., and Soricut, R. (2019). ALBERT: A lite BERT for Self-supervised Learning of Language Representations. *arXiv preprint arXiv:1909.11942*.

Landman, T. and Silverman, B. (2019). Globalization and Modern Slavery. *Politics and Governance*, 7(4):275–290.

Lei, T., Barzilay, R., and Jaakkola, T. (2016). Rationalizing Neural Predictions. *arXiv preprint arXiv:1606.04155*.

LightTag. (2018). The text annotation tool for teams. https://www.lighttag.io/.

Liu, Y., Ott, M., Goyal, N., Du, J., Joshi, M., Chen, D., Levy, O., Lewis, M., Zettlemoyer, L., and Stoyanov, V. (2019). RoBERTa: A Robustly Optimized BERT Pretraining Approach. *arXiv preprint arXiv:1907.11692*.

Lucas, B. and Landman, T. (2021). Social Listening, Modern Slavery, and COVID-19. *Journal of Risk Research*, 24(3-4):314–334.

Madjarov, G., Kocev, D., Gjorgjevikj, D., and Džeroski, S. (2012). An extensive experimental comparison of methods for multi-label learning. *Pattern recognition*, 45(9):3084–3104.

Milivojevic, S., Moore, H., and Segrave, M. (2020). Freeing the Modern Slaves, One Click at a Time: Theorising Human Trafficking, Modern Slavery, and Technology. *Anti-trafficking review*, 45(14):16–32.

Nomnga, P. and Ngqulu, N. (2021). Advancing Artificial Intelligence to Combat Escalating Cyberspace Human Rights Violations in Africa. In *2021 IST-Africa Conference (IST-Africa)*, pages 1–9. IEEE.

Nowak, S. and Rüger, S. (2010). How reliable are annotations via crowdsourcing: A study about inter-annotator agreement for multi-label image annotation. In *Proceedings of the international conference on Multimedia information retrieval*, pages 557–566.

Pasley, R. (2018). Digitalization of Crime Detection in the Supply Chain. *on Logistics (ISL 2018) Big Data Enabled Supply Chain Innovations*.

Rajapakse, T. C. (2019). Simple transformers. https://github.com/ThilinaRajapakse/simpletransformers.

Sanh, V., Debut, L., Chaumond, J., and Wolf, T. (2019). DistilBERT, a distilled version of BERT: smaller, faster, cheaper and lighter. *arXiv preprint arXiv:1910.01108*.

Schapire, R. E. and Singer, Y. (2000). BoosTexter: A Boosting-based System for Text Categorization. *Machine learning*, 39(2):135–168.

Strout, J., Zhang, Y., and Mooney, R. J. (2019). Do Human Rationales Improve Machine Explanations? *arXiv preprint arXiv:1905.13714*.

TAH. (2012). Traffikanalysis.org. `https://www.traffikanalysis.org/`.

Tahir, M. A., Kittler, J., and Yan, F. (2012). Inverse random under sampling for class imbalance problem and its application to multi-label classification. *Pattern Recognition*, 45(10):3738–3750.

Vijayan, V. K., Bindu, K., and Parameswaran, L. (2017). A Comprehensive Study of Text Classification Algorithms. In *2017 International Conference on Advances in Computing, Communications and Informatics (ICACCI)*, pages 1109–1113. IEEE.

Wolf, T., Debut, L., Sanh, V., Chaumond, J., Delangue, C., Moi, A., Cistac, P., Rault, T., Louf, R., and Funtowicz, M. (2019). Huggingface's Transformers: State-of-the-art Natural Language Processing. *arXiv preprint arXiv:1910.03771*.

Yang, Z., Dai, Z., Yang, Y., Carbonell, J., Salakhutdinov, R. R., and Le, Q. V. (2019). XLNet: Generalized Autoregressive Pretraining for Language Understanding. *Advances in neural information processing systems*, 32.

Zaidan, O., Eisner, J., and Piatko, C. (2007). Using "Annotator Rationales" to Improve Machine Learning for Text Categorization. In *Human language technologies 2007: The conference of the North American chapter of the Association for Computational Linguistics; Proceedings of the main conference*, pages 260–267.

## A.  Annotation guidelines

Here are the annotation guidelines shared with our annotators to facilitate the labelling task. The resources are available in LightTag and were shared electronically with the annotation team.

**Overview**

Thank you for agreeing to help us with this task—the annotation of forced labour indicators in text data. We will present you with one news article at a time and ask you to assign forced labour indicators to and tag specific phrases in each one of them. We will be using this data to build a computational model that can recognise risks of forced labour on text data and generate human-understandable justifications for its predictions.

**Instructions**

We would like you to assign indicators of forced labour as defined by the International Labour Organisation (ILO) to news articles. These indicators represent the most common signs or "cues" that point to the possible existence of a forced labour case. They are derived from the theoretical and practical experience of the ILO's Special Action Programme to Combat Forced Labour (SPA-FL). For more information, please **visit**.

We have created a **short video** (less than 5 minutes) describing the tool and the annotation process to facilitate your work. In summary, we are asking you to identify the risks of forced labour in news articles. To assign an ILO indicator to a news article, tag what phrases or sentences led you to decide the presence of that indicator. You can do this by clicking on the label corresponding to the indicator or using the shortcut keys we have defined for you and highlighting the phrases/sentences that support your decision.

**ILO Indicators**

For each news article, please choose one or more of the following tags:

**01.  Abuse of vulnerability:** Referring to people who lack knowledge of the local language or laws, have few livelihood options, belong to a minority religious or ethnic group, have a disability or have other characteristics that set them apart from the majority population.
**Example:** A Chinese maid who worked 365 days a year did not speak a word of French except "good morning" and "good evening".

**02. Abusive working and living conditions:** Forced labour victims may endure living and working in conditions that workers would never freely accept. Work may be performed under conditions that are degrading or hazardous and in severe breach of labour law.
**Example:** "The workers were housed in plastic shacks, drinking contaminated water, and they were kept in holes behind bushes in order to hide them until we left."

**03. Debt bondage:** Victims of forced labour may be working to pay off an incurred or sometimes even inherited debt. The debt can arise from wage advances or loans to cover recruitment or transport costs or from daily living or emergency expenses.
**Example:** "A worker borrowed Rs. 20,000 from a middleman. When he had paid back all but Rs. 4000, the middleman falsely claimed that the worker owed him Rs. 40,000."

**04. Deception:** Deception relates to the failure to deliver what has been promised to the worker, either verbally or in writing. Deceptive practices can include false promises regarding work conditions and wages, the type of work, housing and living conditions, or the employer's identity.
**Example:** "It was my auntie who promised to pay for my school expenses but did not fulfil her promises. Instead, she turned me into a maid."

**05. Excessive overtime:** Referring to the obligation of working excessive hours or days beyond the limits prescribed by national law or collective agreement.
**Example:** "I had to work 19 hours a day without any rest and overtime payment or holiday."

**06. Intimidation and threats:** In addition to threats of physical violence, other common threats used against workers include denunciation to the immigration authorities, loss of wages or access to housing or land, sacking family members, and further worsening of working conditions.
**Example:** "When I told the woman I was working for that I wanted to leave, she threatened me and said that unless I paid $600, she would go to the police and tell them I had no papers."

**07. Isolation:** Workers may not know where they are, the worksite may be far from habitation, and there may be no means of transportation available. But equally, workers may be isolated even within populated areas by being kept behind closed doors or confiscating their mobile phones to prevent them from contacting their families and seeking help.
**Example:** "The camp was in an area that was very difficult to reach. To travel to an urban centre, you had to plan the journey several days in advance."

**08. Physical and sexual violence:** Violence can include forcing workers to take drugs or alcohol to have greater control over them. Violence can also be used to force a worker to undertake tasks that were not part of the initial agreement.
**Example:** "I was regularly slapped, whipped and punched."

**09. Restriction of movement:** Victims of forced labour may be locked up and guarded to prevent them from escaping, at work or while being transported. If workers are not free to enter and exit the work premises, subject to certain restrictions which are considered reasonable, this represents a strong indicator of forced labour.
**Example:** "There were bars on the windows and an iron door, like a prison. It was impossible to escape, not even worth contemplating."

**10. Retention of identity documents:** Referring to the retention by the employer of identity documents or other valuable possessions.
**Example:** "As I passed through immigration, the driver grabbed my passport. I cannot leave because my passport is with the employer, and I cannot move around without it."

**11. Withholding of wages:** Workers may be obliged to remain with an abusive employer while waiting for the wages owed to them.
**Example:** "At the beginning, he promised me a salary and I started to work. He gave me food and sometimes bought me some clothes. But I was still waiting for my salary."

To check a summary of these indicators that might facilitate your work, please refer to our **Cheat sheet** with information about definitions, examples and shortcut keys.

**Additional Instructions**

We have gathered a list of "special" situations that you might encounter while annotating a news article and our recommendation on how to deal with them:

- **Phrases or sentences instead of individual words:** We strongly recommend you tag phrases rather than specific words when assigning an indicator to a news article. In this way, our model can better understand the context in which these indicators appeared.

- **There is no indicator:** If you consider that a particular news article does not contain any indicator of forced labour, submit it and continue to the next one.

- **Two or more phrases/sentences justify my decision:** Your decision of assigning an indicator might be based on more than one phrase/sentence. Please highlight all phrases/sentences relevant to your decision (there is no limit on the number of phrases/sentences that can support your decision).

- **There is a phrase/sentence that supports my decision for two or more indicators:** Unfortunately, LightTag allows the use of a sentence or phrase as justification for just one indicator. Consequently, please highlight the sentence/phrase with the risk you consider is more strongly related.

If there is any other situation not covered in these guidelines, or if you have suggestions on how to improve them, please reach out to the research team.
Many thanks!

## B. Inter-annotator agreement

Table 5 illustrates the inter-annotator agreement (F1 score) for each forced labour indicator both at a label and rationale level.

| Label | F1 Score - Labels | F1 Score - Rationales |
|---|---|---|
| Abuse of vulnerability | 0.81 | 0.72 |
| Abusive working and living conditions | 0.86 | 0.78 |
| Debt bondage | 0.87 | 0.78 |
| Deception | 0.65 | 0.59 |
| Excessive overtime | 0.76 | 0.66 |
| Intimidation and threats | 0.74 | 0.67 |
| Isolation | 0.71 | 0.63 |
| Physical and sexual violence | 0.82 | 0.74 |
| Restriction of movement | 0.81 | 0.72 |
| Retention of identity documents | 0.84 | 0.76 |
| Withholding of wages | 0.92 | 0.83 |

Table 5: Per-class F1 scores for Inter-Annotation Agreement

## C. Hyperparameter tuning

Here are the details of the hyperparameter tuning process for the classification experiments.

Table 6 describes the search space for each hyperparameter in terms of their sampling distribution and possible values. As mentioned in our paper, these values are tuned for each classifier using a random search method.

| Hyperparameter | Distribution | Value ranges |
|---|---|---|
| learning rate | log uniform | $[log(0.00001), log(0.01)]$ |
| threshold | random | [0.15, 0.20, 0.25, 0.30, 0.35, 0.40] |
| train batch size | random | [2, 4, 6, 8] |
| optimizer | random | ['AdamW', 'Adafactor'] |

Table 6: Hyperparameter search space

Finally, Table 7 shows each classifier's hyperparameter values for their fine-tuning.

| Model | LR | Threshold | Batch Size | Optimiser |
|---|---|---|---|---|
| distilbert-base | $2.61 \times 10^{-3}$ | 0.20 | 2 | Adafactor |
| albert-base | $1.74 \times 10^{-3}$ | 0.25 | 2 | Adafactor |
| roberta-base | $1 \times 10^{-3}$ | 0.20 | 2 | Adafactor |
| distilroberta-base | $2.34 \times 10^{-5}$ | 0.30 | 2 | AdamW |
| roberta-large | $9.66 \times 10^{-4}$ | 0.40 | 2 | Adafactor |
| xlnet-base | $2.68 \times 10^{-5}$ | 0.25 | 4 | AdamW |

Table 7: Hyperparameters used for fine-tuning

## D. Detailed classification results

This section details the performance evaluation for each classifier.

| Metric | Sample 1 | Sample 2 | Sample 3 |
|---|---|---|---|
| F1(weighted) | 0.47 | 0.45 | 0.40 |
| F1(micro) | 0.45 | 0.47 | 0.39 |
| F1(macro) | 0.41 | 0.36 | 0.37 |
| LRAP | 0.86 | 0.87 | 0.88 |
| EMR | 0.49 | 0.42 | 0.05 |

Table 8: Performance metrics for roberta-base

| Label | Sample 1 | Sample 2 | Sample 3 |
|---|---|---|---|
| Abuse of vulnerability | 0.25 | 0.48 | 0.31 |
| Abusive working and living conditions | 0.70 | 0.56 | 0.49 |
| Debt bondage | 0.50 | 0.41 | 0.22 |
| Deception | 0.25 | 0.33 | 0.44 |
| Excessive overtime | 0.54 | 0.50 | 0.43 |
| Intimidation and threats | 0.21 | 0.45 | 0.45 |
| Isolation | 0.45 | 0.24 | 0.36 |
| Physical and sexual violence | 0.35 | 0.48 | 0.38 |
| Restriction of movement | 0.47 | 0.00 | 0.38 |
| Retention of identity documents | 0.15 | 0.18 | 0.20 |
| Withholding of wages | 0.61 | 0.29 | 0.44 |

Table 9: Per-class F1 scores for roberta-base

| Metric | Sample 1 | Sample 2 | Sample 3 |
|---|---|---|---|
| F1(weighted) | 0.49 | 0.50 | 0.43 |
| F1(micro) | 0.49 | 0.49 | 0.42 |
| F1(macro) | 0.45 | 0.47 | 0.42 |
| LRAP | 0.85 | 0.88 | 0.89 |
| EMR | 0.50 | 0.43 | 0.09 |

Table 10: Performance metrics for distilroberta-base

| Label | Sample 1 | Sample 2 | Sample 3 |
|---|---|---|---|
| Abuse of vulnerability | 0.20 | 0.31 | 0.28 |
| Abusive working and living conditions | 0.68 | 0.66 | 0.54 |
| Debt bondage | 0.28 | 0.24 | 0.15 |
| Deception | 0.46 | 0.63 | 0.54 |
| Excessive overtime | 0.58 | 0.57 | 0.44 |
| Intimidation and threats | 0.36 | 0.62 | 0.67 |
| Isolation | 0.50 | 0.33 | 0.43 |
| Physical and sexual violence | 0.51 | 0.52 | 0.39 |
| Restriction of movement | 0.57 | 0.55 | 0.62 |
| Retention of identity documents | 0.25 | 0.25 | 0.18 |
| Withholding of wages | 0.60 | 0.50 | 0.43 |

Table 11: Per-class F1 scores for distilroberta-base

| Metric | Sample 1 | Sample 2 | Sample 3 |
|---|---|---|---|
| F1(weighted) | 0.49 | 0.44 | 0.36 |
| F1(micro) | 0.48 | 0.44 | 0.37 |
| F1(macro) | 0.47 | 0.39 | 0.35 |
| LRAP | 0.86 | 0.88 | 0.88 |
| EMR | 0.48 | 0.25 | 0.06 |

Table 12: Performance metrics for distilbert-base

| Label | Sample 1 | Sample 2 | Sample 3 |
|---|---|---|---|
| Abuse of vulnerability | 0.30 | 0.36 | 0.28 |
| Abusive working and living conditions | 0.65 | 0.60 | 0.44 |
| Debt bondage | 0.35 | 0.24 | 0.24 |
| Deception | 0.50 | 0.33 | 0.36 |
| Excessive overtime | 0.61 | 0.51 | 0.41 |
| Intimidation and threats | 0.16 | 0.52 | 0.36 |
| Isolation | 0.55 | 0.22 | 0.29 |
| Physical and sexual violence | 0.34 | 0.48 | 0.39 |
| Restriction of movement | 0.75 | 0.55 | 0.55 |
| Retention of identity documents | 0.22 | 0.25 | 0.22 |
| Withholding of wages | 0.72 | 0.27 | 0.32 |

Table 13: Per-class F1 scores for distilbert-base

| Metric | Sample 1 | Sample 2 | Sample 3 |
|---|---|---|---|
| F1(weighted) | 0.51 | 0.44 | 0.38 |
| F1(micro) | 0.52 | 0.45 | 0.37 |
| F1(macro) | 0.47 | 0.41 | 0.38 |
| LRAP | 0.87 | 0.86 | 0.87 |
| EMR | 0.51 | 0.43 | 0.06 |

Table 14: Performance metrics for xlnet-base

| Label | Sample 1 | Sample 2 | Sample 3 |
|---|---|---|---|
| Abuse of vulnerability | 0.34 | 0.19 | 0.19 |
| Abusive working and living conditions | 0.68 | 0.61 | 0.49 |
| Debt bondage | 0.50 | 0.35 | 0.24 |
| Deception | 0.50 | 0.50 | 0.50 |
| Excessive overtime | 0.58 | 0.50 | 0.45 |
| Intimidation and threats | 0.20 | 0.58 | 0.54 |
| Isolation | 0.55 | 0.20 | 0.19 |
| Physical and sexual violence | 0.38 | 0.57 | 0.44 |
| Restriction of movement | 0.75 | 0.25 | 0.67 |
| Retention of identity documents | 0.22 | 0.25 | 0.29 |
| Withholding of wages | 0.54 | 0.46 | 0.18 |

Table 15: Per-class F1 scores for xlnet-base

| Metric | Sample 1 | Sample 2 | Sample 3 |
|---|---|---|---|
| F1(weighted) | 0.47 | 0.47 | 0.35 |
| F1(micro) | 0.47 | 0.47 | 0.34 |
| F1(macro) | 0.46 | 0.44 | 0.31 |
| LRAP | 0.86 | 0.88 | 0.87 |
| EMR | 0.44 | 0.34 | 0.04 |

Table 16: Performance metrics for albert-base

| Label | Sample 1 | Sample 2 | Sample 3 |
|---|---|---|---|
| Abuse of vulnerability | 0.34 | 0.37 | 0.23 |
| Abusive working and living conditions | 0.59 | 0.59 | 0.50 |
| Debt bondage | 0.25 | 0.15 | 0.14 |
| Deception | 0.66 | 0.53 | 0.28 |
| Excessive overtime | 0.56 | 0.50 | 0.27 |
| Intimidation and threats | 0.22 | 0.57 | 0.47 |
| Isolation | 0.66 | 0.35 | 0.25 |
| Physical and sexual violence | 0.34 | 0.52 | 0.41 |
| Restriction of movement | 0.75 | 0.60 | 0.22 |
| Retention of identity documents | 0.22 | 0.25 | 0.33 |
| Withholding of wages | 0.44 | 0.36 | 0.32 |

Table 17: Per-class F1 scores for albert-base

| Metric | Sample 1 | Sample 2 | Sample 3 |
|---|---|---|---|
| F1(weighted) | 0.47 | 0.46 | 0.39 |
| F1(micro) | 0.47 | 0.46 | 0.40 |
| F1(macro) | 0.44 | 0.44 | 0.37 |
| LRAP | 0.86 | 0.87 | 0.88 |
| EMR | 0.55 | 0.49 | 0.13 |

Table 18: Performance metrics for roberta-large

| Label | Sample 1 | Sample 2 | Sample 3 |
|---|---|---|---|
| Abuse of vulnerability | 0.27 | 0.30 | 0.32 |
| Abusive working and living conditions | 0.63 | 0.62 | 0.51 |
| Debt bondage | 0.41 | 0.29 | 0.29 |
| Deception | 0.50 | 0.42 | 0.26 |
| Excessive overtime | 0.60 | 0.38 | 0.28 |
| Intimidation and threats | 0.20 | 0.61 | 0.52 |
| Isolation | 0.53 | 0.25 | 0.21 |
| Physical and sexual violence | 0.38 | 0.57 | 0.52 |
| Restriction of movement | 0.60 | 0.46 | 0.55 |
| Retention of identity documents | 0.25 | 0.44 | 0.50 |
| Withholding of wages | 0.44 | 0.46 | 0.15 |

Table 19: Per-class F1 scores for roberta-large